%% file: main.tex
\documentclass[sigconf]{acmart}

\AtBeginDocument{%
  }

\usepackage[utf8]{inputenc} % allow utf-8 input
\usepackage{hyperref}       % hyperlinks
\usepackage{url}            % simple URL typesetting
\usepackage{booktabs}       % professional-quality tables
\usepackage{amsfonts}       % blackboard math symbols
\usepackage{nicefrac}       % compact symbols for 1/2, etc.
\usepackage{microtype}      % microtypography
\usepackage{xcolor}         % colors
\usepackage{graphicx}
\usepackage{import}
\usepackage{xspace}
\usepackage{fancyhdr}

\usepackage{macros}

\renewcommand\footnotetextcopyrightpermission[1]{}

\begin{document}

\title{Automatic Measures for Evaluating\\ Generative Design Methods for Architects}


\author{Eric Yeh}
\affiliation{
    \institution{Artificial Intelligence Center\\SRI International}
    \city{Menlo Park}
    \state{CA}
    \country{USA}
}
\email{eric.yeh@sri.com}

\author{Briland Hitaj}
\affiliation{
    \institution{Computer Science Laboratory\\SRI International}
    \city{New York}
    \state{NY}
    \country{USA}
}
\email{briland.hitaj@sri.com}

\author{Vidyasagar Sadhu}
\affiliation{
    \institution{Artificial Intelligence Center\\SRI International}
    \city{Menlo Park}
    \state{CA}
    \country{USA}
}
\email{srikanthvidyasagar.sadhu@sri.com}

\author{Anirban Roy}
\affiliation{
    \institution{Computer Science Laboratory\\SRI International}
    \city{Menlo Park}
    \state{CA}
    \country{USA}
}
\email{anirban.roy@sri.com}

\author{Takuma Nakabayashi}
\affiliation{
    \institution{Obayashi Corporation}
    \city{Menlo Park}
    \state{CA}
    \country{USA}
}
\email{takuma.nakabayashi@obayashi-usa.com}

\author{Yoshito Tsuji}
\affiliation{
    \institution{Obayashi Corporation}
    \city{Tokyo}
    \country{Japan}
}
\email{tsuji.yoshito@obayashi.co.jp}

\renewcommand{\shortauthors}{Yeh et al.}

\import{\sectiondir}{abstract.tex}



%% Excised for now
\begin{teaserfigure}
  \centering
  \includegraphics[width=0.75\textwidth]{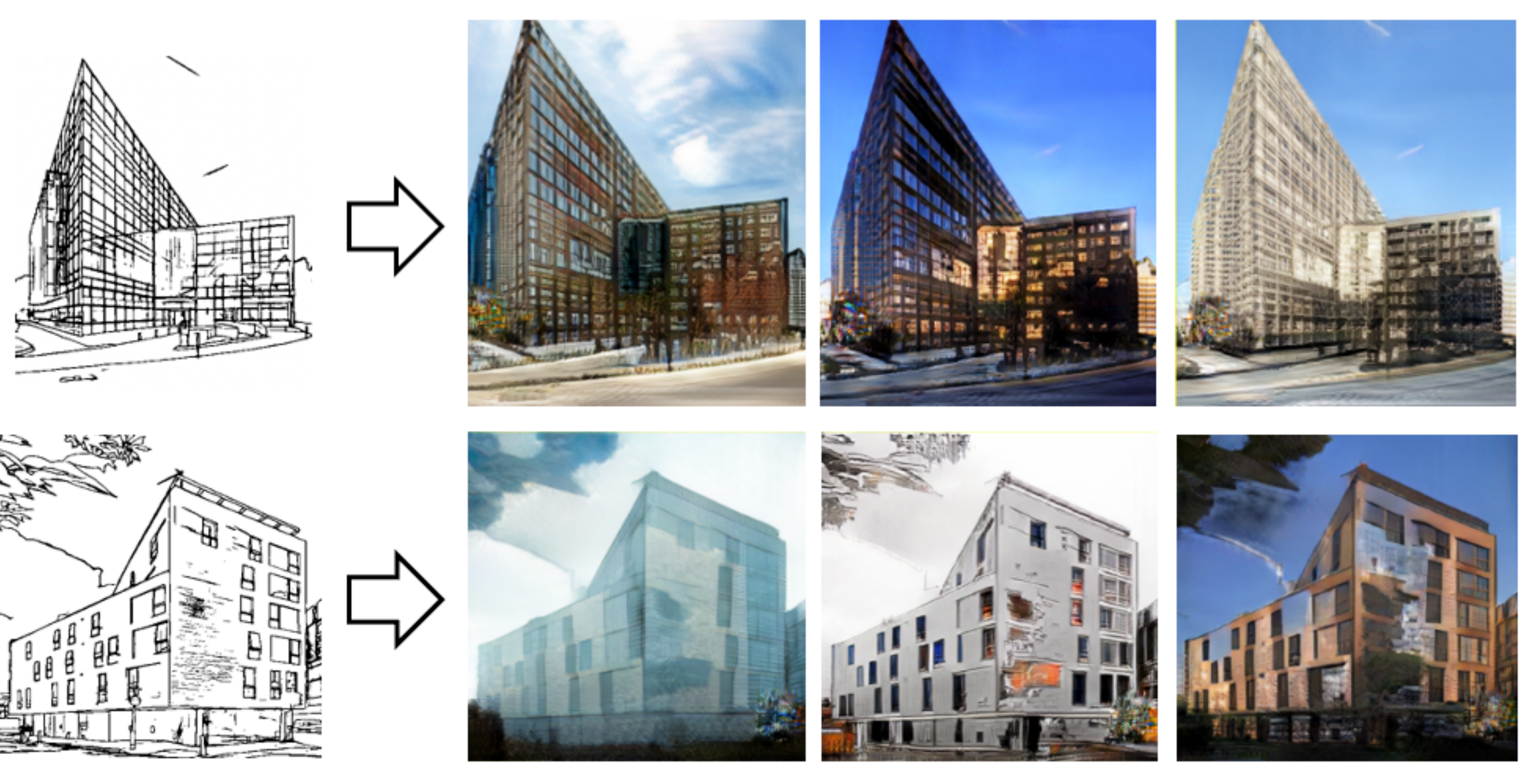}
  \caption{Examples of generated building candidates that meet our identified architectural criteria of realism, matching intent, and structural diversity. Our proposed generative models take conceptual sketches (left) and create a diverse set of realistic architectural designs (right) that match the content and the overall structural intent of the sketch, while exploring a variety of lower-level structural and material differences. }
  %\caption{Our aim is to identify generative models that can take conceptual sketches (left) and create a diverse set of realistic architectural designs (right) that match the content, the overall structural intent of the sketch, while exploring a variety of lower-level structural and material differences.}
  \Description{Examples of generated building candidates that meet our identified architectural criteria of realism, matching intent, and structural diversity. Our proposed generative models take conceptual sketches (left) and create a diverse set of realistic architectural designs (right) that match the content and the overall structural intent of the sketch, while exploring a variety of lower-level structural and material differences.  }
  \label{fig:teaser}
\end{teaserfigure}

%

\maketitle
\pagestyle{fancy}
\fancyhead{}
\fancyfoot{}
\fancyhead[L]{\footnotesize Automatic Measures for Evaluating Generative Design Methods for Architects}
\fancyhead[R]{\footnotesize Yeh et al.}
\fancyfoot[C]{\thepage}

\import{\sectiondir}{introduction.tex}

\import{\sectiondir}{approach.tex}
\import{\sectiondir}{experiments.tex}
\import{\sectiondir}{conclusions.tex}
\import{\sectiondir}{acknowledgments.tex}

\bibliographystyle{ACM-Reference-Format}
\bibliography{main}

\end{document}

%% file: sections/abstract.tex
% !TEX root = ../main.tex
\begin{abstract}
%% What are the key ideas and takeaways readers should know in this abstract?
The recent explosion of high-quality image-to-image methods has prompted interest in applying image-to-image methods towards artistic and design tasks.
Of interest for architects is to use these methods to generate design proposals from conceptual sketches, usually hand-drawn sketches that are quickly developed and can embody a design intent.
More specifically, instantiating a sketch into a visual that can be used to elicit client feedback is typically a time consuming task, and being able to speed up this iteration time is important.
While the body of work in generative methods has been impressive, there has been a mismatch between the quality measures used to evaluate the outputs of these systems and the actual expectations of architects.
In particular, most recent image-based works place an emphasis on realism of generated images.
While important, this is one of several criteria architects look for.
In this work, we describe the expectations architects have for design proposals from conceptual sketches, and identify corresponding automated metrics from the literature.
We then evaluate several image-to-image generative methods that may address these criteria and examine their performance across these metrics.
From these results, we identify certain challenges with hand-drawn conceptual sketches and describe possible future avenues of investigation to address them.
\end{abstract}

%% file: sections/introduction.tex
% !TEX root = ../main.tex
\section{Introduction}
\label{sec:intro}

%% What are we doing and why is it important?
%% What are our primary scientific claims?
%% How do answering our claims contribute to the importance of the tool?

With the rise of autoregressive and iterative generative models and large available datasets, the quality of generated images has gone up dramatically along with interest in applying these models to real tasks.  
In this work, we study how current generative methods can facilitate exploration of the architectural design space.
While the initial design process varies from person to person, we focus on the common strategy of creating realistic and usable architectural renderings from conceptual sketches.

 \begin{figure}[h]
\begin{center}
\includegraphics[width=.55\columnwidth]{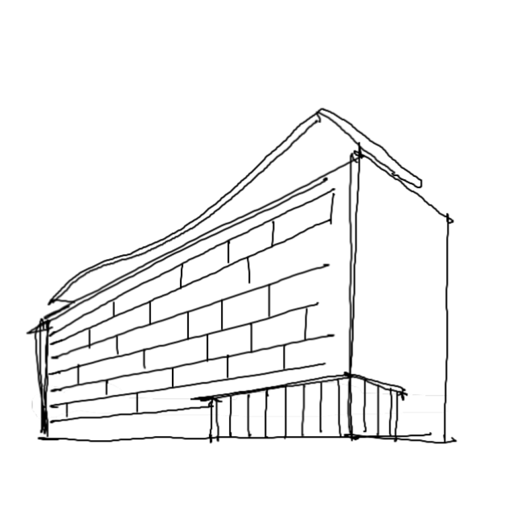}
\end{center}
\caption{A sample conceptual sketch.}
\label{fig:conceptual_sketch}
\end{figure}

Figure \ref{fig:conceptual_sketch} gives an example of a conceptual sketch used in the initial design phase of a project.
Sketches are meant to quickly explore the design space, and as a result they exhibit certain characteristics. 
These sketches are hand-drawn, often sparsely populated, as they are neither intended to be exacting nor definitive.
Indeed, lines and markings made are often not intended to be followed strictly, as with edge maps, and instead use drawn motifs to convey the essence, the broad intents about what should be present.
For example, the horizontal line markings on the left-side facade may be interpreted as a series of wide windows. 
How this will be realized, whether these will be met with large windows with small insets, and the type of material used (e.g., steel, concrete, or other materials), are left unspecified. 

From our interviews with the architectural design team at Obayashi Corporation, we have identified three main criteria for renders: 

\begin{itemize}
    \item \textbf{Overall Intent}: Generated renders should contain the same overall structure as communicated by the sketch.
    \item \textbf{Realism}: Renders should be appear to be realistic buildings, with no visual artifacts.
    \item \textbf{Structural Variety}: Renders should contain a variety of structural and material elements while still respecting intended structure in the sketch.
\end{itemize}

For the rest of this work, we first describe our automated proxies for these criteria.  We then detail the image-to-image generative methods used to generate candidate designs from a corpus of conceptual sketches, with training details where relevant.  An evaluation against the automated measures is performed.  We follow with a discussion of these results and highlight issues with using conceptual sketches as inputs and describe future work to address these issues.

%% 

%% file: sections/approach.tex
% !TEX root = ../main.tex
\section{Approach}
\label{sec:approach}

\begin{table*}[]
 \caption{Evaluation of different sketch to image methods meeting content distance from the conceptual sketch (lower is better), similarity to real images (FID, lower is better), and structural diversity of designs (higher is better). }
     \centering
     \begin{tabular}{c|c|c|c}
         \textbf{Method} & \textbf{Content Distance $\downarrow$} & \textbf{FID $\downarrow$} & \textbf{Structural Diversity $\uparrow$} \\
         \hline 
          MUNIT &  $0.43 \pm 0.15 $ & $188.05$ & $ 0.37 \pm 0.22 $ \\
          Pix2pix & $ 0.54 \pm 0.08$ & $294.23$ & $0.0 \pm 0.0$ \\
          ControlNet & $ 0.54 \pm 0.12$ & $190.48$ & $ 0.80 \pm 0.36$ \\
          Conceptual Sketches &  $\mathbf{0.07 \pm 0.02}$ & $357.58$ & $ 0.0 \pm 0.0$\\
          \hline
     \end{tabular}
     
     \label{tab:experiment}
 \end{table*}

In our work, we have identified several evaluation automated measures that can act as proxies for the generated renders desiderata:
\begin{itemize}
    \item \textbf{Content distance}: L1 distance between the MUNIT \cite{munit_ref} content encoder of the sketch and the rendered designs;
    \item \textbf{Frechet Inception Distance} (FID) \cite{FID_ref} to measure the realism of the candidate designs;
    \item \textbf{Structural diversity}, computed using Structural Similarity Index Measure (SSIM) \cite{SSIM_ref}.
\end{itemize}

\noindent\textbf{Content distance} is our proxy for determining if the overall intent of a generated design matches those of a sketch.
To measure this, we use Multimodal Unsupervised Image-to-Image Translation (MUNIT) which is an autoencoder model that uses cyclic consistency to generate neural encoders for the style and content of an image \cite{neural_style_ref}.  Style refers to the distribution of pixels within small patches of an image.  Content refers to the broader structure of an image, and is invariant across application of multiple different styles.  For example, an image of a building painted blue will have a different style compared to the same building being painted red, but have similar contents.  This also extends to lower level structural details that may be captured by the style coding.  
MUNIT employs a cyclic loss \cite{cyclegan_ref} to translate between the content of an image and the variety of different renderings that can stem from it.  
In our case, we use MUNIT to learn how to identify the content from a sketch and how it applies to different building images.

 \begin{figure}[h]
\begin{center}
\includegraphics[width=1.\columnwidth]{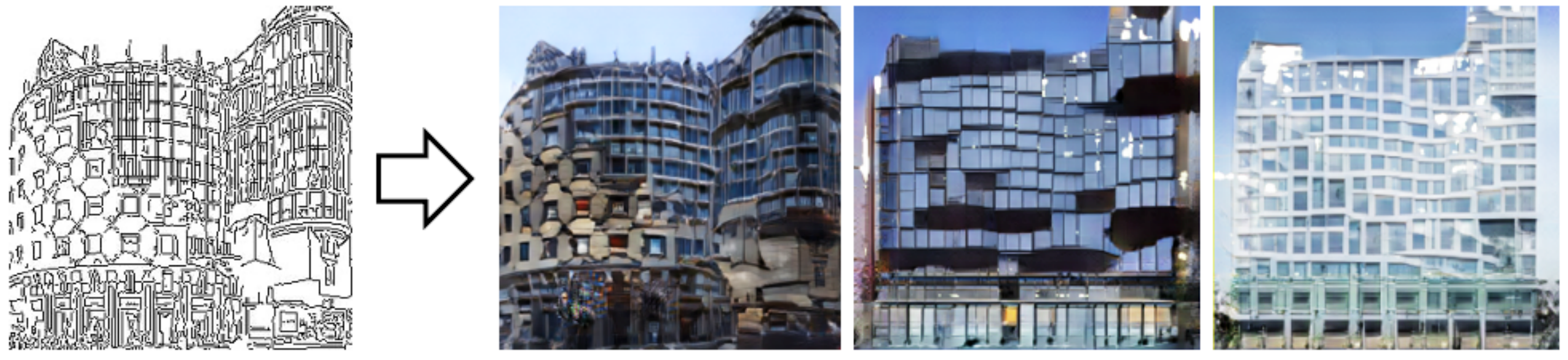}
\end{center}
\caption{Capturing content versus style: The content, broad structural information, of the original sketch is combined with multiple styles encoding different low-level pixel information to produce several visually different results.}
\label{fig:munit_content}
\end{figure}

Figure \ref{fig:munit_content} illustrates how content is preserved across multiple different renders.  Here a trained MUNIT network creates three visually distinct renders from a single source sketch. 

\noindent\textbf{Frechet Inception Distance (FID)} is the predominant measure used to measure the fidelity of generated images with a corpus of in-domain real images.  This is done by computing from the feature activation statistics of an Inception network \cite{inception_ref} over the corpus of real images and synthetic images.  The Frechet divergence (the squared Wasserstein distance) between the distribution of visual features between real and synthetic images is used to determine how unrealistic a given generative method is.

\noindent\textbf{Structural diversity} is taken as the inverse of the Structural Similarity Index Measure (SSIM) \cite{SSIM_ref}. SSIM specifically aims at capturing structural relationships, where structure is represented by spatially close pixels that are also strongly correlated.  For each set of designs generated from a conceptual sketch, we take the mean of the pairwise SSIMs for all renders from a given conceptual sketch. 

Inception score \cite{inception_ref} is another popular measure used to assess the quality and diversity of generated images.  However, this score is computed from feature activation statistics in an Inception network, and features corresponding to palette and low-level texture changes may be captured and treated as indicating diversity.  The stronger assumptions about structure in SSIM thus are a better match.
We explore the use of several image-to-image models for our design goals.
These models are: 1) Pix2pix \cite{pix2pix2017}, 2) MUNIT \cite{munit_ref}, and 3) ControlNet \cite{controlnet_ref}.
% \begin{itemize}
%     \item Pix2pix \cite{pix2pix2017}
%     \item MUNIT \cite{munit_ref}
%     \item ControlNet \cite{controlnet_ref}
% \end{itemize}

Pix2pix is a framework that uses conditional generative adversarial networks to learn transformations from one domain of images to another.  
Both Pix2pix and MUNIT were trained using a dataset of $80,000$ architectural images drawn from Flickr\footnote{https://www.flickr.com/} and Unsplash\footnote{https://unsplash.com/}.  
ControlNet is a state-of-the art method that uses sketches as guidance cues for controlling the inference in a pre-trained diffusion model.  By forcing the generated images match the constraints provided by the sketches, ControlNet can leverage existing highly-trained diffusion models without having to fine-tune or retrain the model.  For these experiments, we used the scribble maps guidance with the prompt "a photograph of a building."
A major challenge is the lack of a large scale corpus of architectural conceptual sketches.  
Previous studies on sketch-to-image generation have synthesized sketch-like images from images using heuristic methods such as Canny Edge Detection \cite{canny_edge_ref} or the Hough transform \cite{hough_ref}.  Neural methods have also been trained to generate edges based on specific criteria.  Commonly used examples of this class include Holistically-Nested Edge Detection (HED) \cite{hed_ref} and DexiNed \cite{dexined_ref}.  For our training models, we used edges detected via DexiNed to generate corresponding sketch images for our data, as these most visually resemble hand-drawn sketches.

%  \begin{figure}[]
% \begin{center}
% \includegraphics[width=\columnwidth]{figures/pix2pix_averaging.pdf}
% \end{center}
% \caption{Averaging effects can be observed from a sample of renders produced by Pix2pix.}
% \label{fig:pix2pix_averaging}
% \end{figure}

%% file: sections/experiments.tex
% !TEX root = ../main.tex
\section{Experiments and Results}
\label{sec:experiements}

As part of our evaluation,
we elicited $24$ conceptual sketches from a team of architects.  
We then sampled $10$ designs from each method.  
For comparison, we also ran the conceptual sketches ``as-is'' across each of the measures.
Table \ref{tab:experiment} shows performance of different methods against each of the criteria.  Unsurprisingly, using conceptual sketches gives the best match with content distance, with a poorer FID (as sketches do not resemble real images).  Because there is no variation in these sketches, the structural diversity is poor as well.

MUNIT exhibited the best content distance, significantly outperforming the other approaches \footnote{Using a two tail t-test with $\alpha=0.05$} while also having a slight loss in realism  (lower FID).  ControlNet does exhibit a higher structural diversity score, albeit at a cost in content distance.

 \begin{figure}[]
\begin{center}
\includegraphics[width=1.\columnwidth]{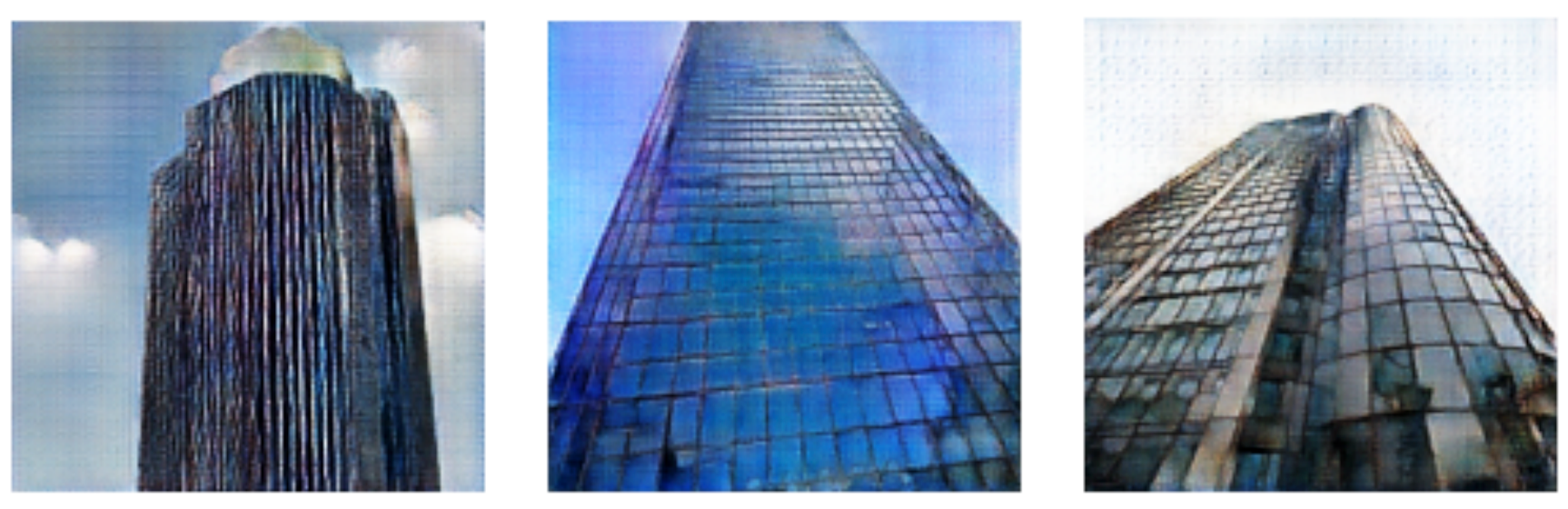}
\end{center}
\caption{Averaging effects can be observed from a sample of renders produced by Pix2pix.}
\label{fig:pix2pix_averaging}
\end{figure}

Pix2pix is a deterministic mapping that only generates one render from a given sketch, and thus has a zero structural diversity score.  Its realism is also poorer as well.  While one could argue this implementation does not have the advanced decoding capabilities seen in the newer networks, a visual inspection of the validation renders showed an averaging effect, such as  colors being muted to a handful of similar palettes (Figure \ref{fig:pix2pix_averaging}).

%%

%% file: sections/conclusions.tex
% !TEX root = ../main.tex
\section{Findings and Discussion}
\label{sec:conclusions}

%% Restate what we did, primary findings
%% What is the future work in this area?
While it may be unsurprising that the network used to derive content distance also performed the best at it, MUNIT also performed reasonably well on the other measures and visual inspection found it displayed a good variety of palette and structural differences. This would argue for approaches that make a distinction between content and style that can be used to implement our criteria.

\begin{figure}[h]
\begin{center}
\includegraphics[width=1\columnwidth]{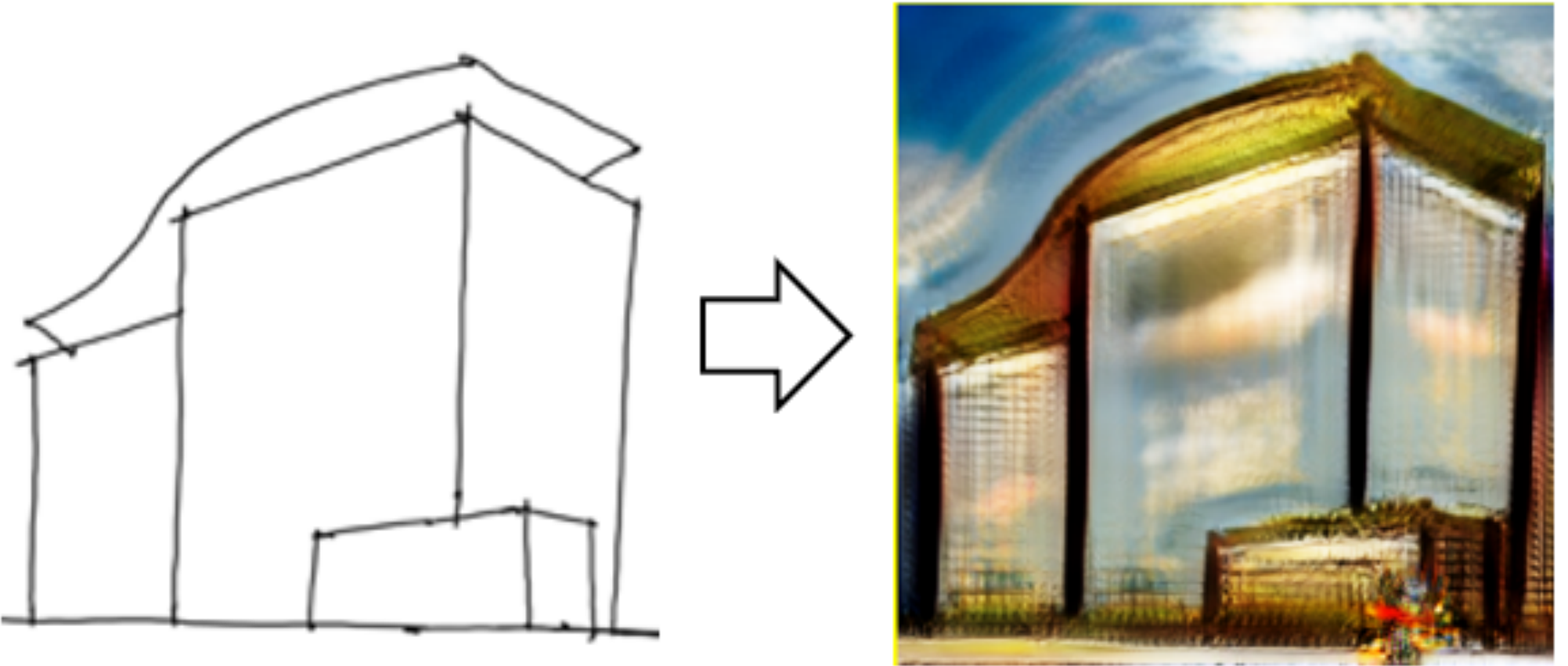}
\end{center}
\caption{The effect of sparsity on image-to-image generative methods.  A sparse conceptual sketch (left) results in poor results using highly trained models such as MUNIT (right).}
\label{fig:sparsity}
\end{figure}

A cursory inspection of validation results overall appeared to give impressive results.  However, the actual performance on conceptual sketches was considerably poorer.
This first issue stems from the comparative sparsity of conceptual sketches.  Most image-to-image frameworks, particularly those that explicitly model and style and content differentiation \cite{cyclegan_ref, munit_ref}, tend to have relatively complex source and target images.  In the case of conceptual sketches, these are informationally impoverished, often resulting in poorer renders.  Figure \ref{fig:sparsity} gives an example of a conceptual sketch (left) being rendered by the MUNIT model (right).

\begin{figure}[]
\begin{center}
\includegraphics[width=\columnwidth]{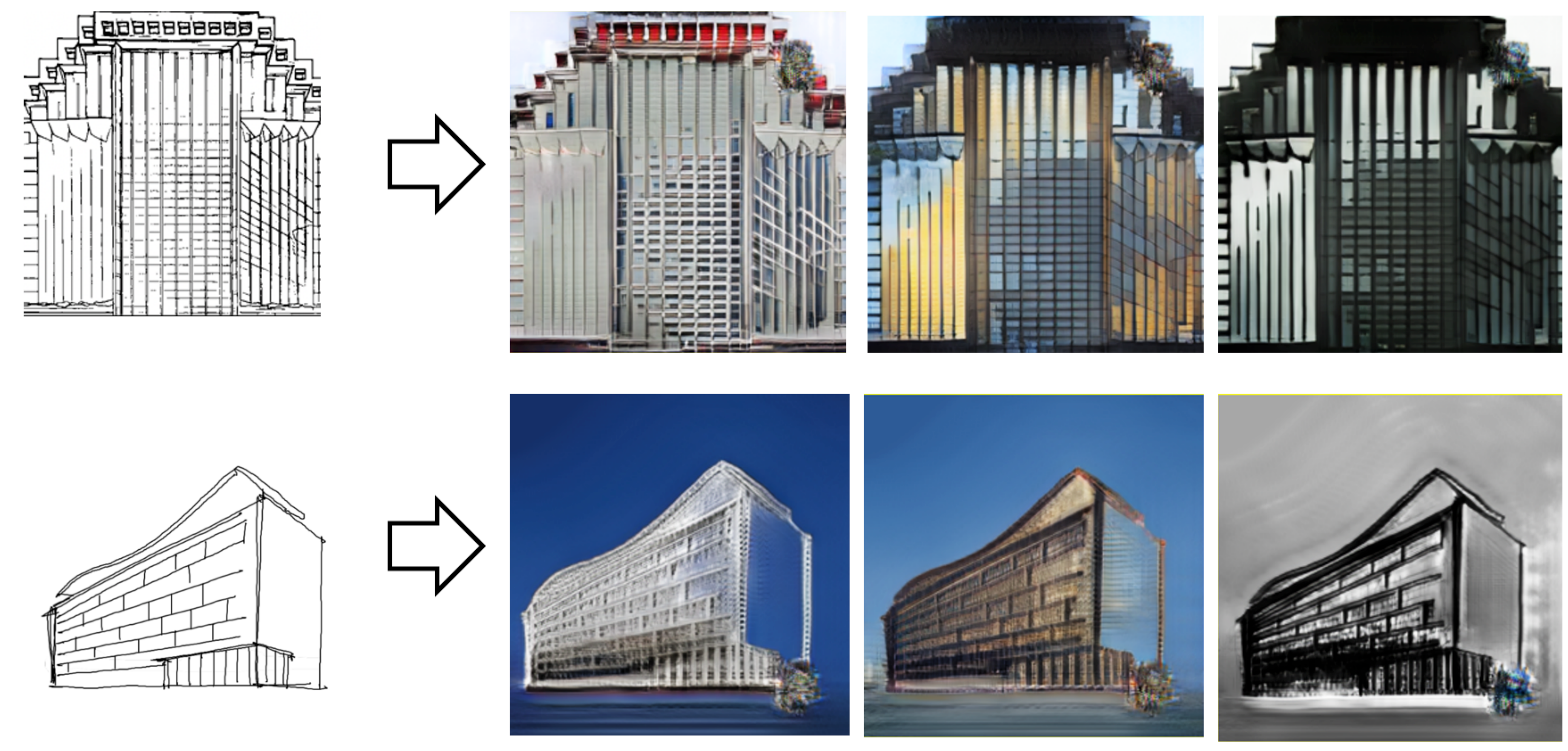}
\end{center}
\caption{Distributional differences between automatically inferred sketches versus hand-drawn sketches.  Renders created by the MUNIT model (top-right) with input from an automatically generated edge map designed to resemble a hand-drawn sketch (top-left) are high quality.  In contrast, when the same model is given a hand-drawn conceptual sketch, the results are less desirable (bottom row). }
\label{fig:idvsood}
\end{figure}

Another major issue lies with distributional mismatches between automatically generated sketch inputs and actual hand-drawn conceptual sketches.  Figure \ref{fig:idvsood} provides an illustrative example using the MUNIT model trained to convert automatically detected edges into a variety of realistic renders.  The top row shows an input sketch and candidate renders generated by the model.  The bottom shows an actual hand-drawn conceptual sketch and corresponding renders using the same model.  While the automatically derived sketch may appear hand-drawn, there are significant enough differences between them that the overall quality of the hand-drawn renders is lower.

Finally, there is a lack of examples from the conceptual sketch domain.  While automatically generated sketches are available, we have found them to be different enough that models trained on them consider hand sketches to be out of distribution inputs.

Future work will need to address this distributional gap between conceptual sketches and automatically generated ones.  The simplest and most correct solution would be to elicit a larger corpus of conceptual sketches, but this would also be the most 
% expensive 
% or costly one. 
expensive given the cost.  
Possible avenues of investigation include the use of image-to-image translation methods for domain adaptation to map between conceptual sketches and artificially generated sketches \cite{img2img_domain_adaptation_ref}.
In addition to the above, we intend to develop human studies to identify correlation between human judgements of intent preservation, realism, and structural variety with our proposed automated measures.
% 

%% file: sections/acknowledgments.tex
% !TEX root = ../main.tex
% \section{Acknowledgments}
% \label{sec:acknowledgments}

%% TODO: Ablate this for anonymized review.
\begin{acks}
This work was supported by a grant from Obayashi Corporation.  The authors would also like to thank the team at the Obayashi Tokyo offices for their feedback and insights into the architectural design process.
\end{acks}